\documentclass{article}
\usepackage{amsmath}
\usepackage{graphicx}

\usepackage[numbers]{natbib}

\usepackage[preprint]{paper}

\usepackage[utf8]{inputenc} 
\usepackage[T1]{fontenc}    
\usepackage{hyperref}       
\usepackage{url}            
\usepackage{booktabs}       
\usepackage{amsfonts}       
\usepackage{nicefrac}       
\usepackage{microtype}      
\usepackage[usenames,dvipsnames,svgnames,table]{xcolor}
\usepackage{subcaption}
\usepackage{footnote}
\makesavenoteenv{tabular}
\makesavenoteenv{table}

\title{Learning to Detect Objects with a 1 Megapixel Event Camera}

\newcommand{\amosSrmk}[1]{{\color{ForestGreen} {\bf as: #1}}}

\newcommand{\JonathanMrmk}[1]{{\color{teal} {\bf jm: #1}}}

\newcommand{\comment}[1]{}
\author{%
   Etienne Perot\\
   PROPHESEE, Paris\\
   \texttt{eperot@prophesee.ai}\\
   \And
   Pierre de Tournemire\\
   PROPHESEE, Paris\\
   \texttt{pdetournemire@prophesee.ai}\\
   \AND
   Davide Nitti\\
   PROPHESEE, Paris\\
   \texttt{dnitti@prophesee.ai}\\
   \And
   Jonathan Masci\\
   NNAISENSE, Lugano\\
   \texttt{jonathan@nnaisense.com}\\
   \And
   Amos Sironi\\
   PROPHESEE, Paris\\
   \texttt{asironi@prophesee.ai}\\
}

\begin{document}

\maketitle

\begin{abstract}
    Event cameras encode visual
    information with high temporal precision, 
    low data-rate, and high-dynamic range. 
    Thanks to these characteristics, event cameras 
    are particularly suited for scenarios with high motion, 
    challenging lighting conditions and requiring low latency. 
    However, due to the novelty of the field, 
    the performance of event-based systems 
    on many vision tasks
    is still lower compared to conventional frame-based solutions.
    The main reasons for this performance gap are: the lower spatial resolution of event sensors, 
    compared to frame cameras; the lack of large-scale training datasets;
    the absence of well established deep learning architectures for event-based processing. 
    In this paper, we address all these problems in the context of an event-based object detection task.
    First, we publicly release the first high-resolution large-scale dataset for object detection.
    The dataset contains more than 14 hours recordings of a 1 megapixel event camera,
    in automotive scenarios, together with 25M bounding boxes of cars, pedestrians, and two-wheelers, 
    labeled at high frequency.
    Second, we introduce a novel recurrent architecture for event-based detection
    and a temporal consistency loss for better-behaved training.
    The ability to compactly represent the sequence of events into the internal memory
    of the model is essential to achieve high accuracy. Our model outperforms by a 
    large margin feed-forward event-based architectures.
    Moreover, our method does not require any reconstruction 
    of intensity images from events, 
    showing that training directly from raw events is possible,
    more efficient, and more accurate than 
    passing through an intermediate intensity image. 
    Experiments on the dataset introduced in this work, for which events
    and gray level images are available, show performance on par with that of
    highly tuned and studied frame-based detectors.
    
    \end{abstract}

\section{Introduction}

 

Event cameras~\cite{Lichtsteiner2008,posch2014retinomorphic,son20174,finateu20205} promise a paradigm shift in computer vision by
representing visual information in a fundamentally different way. 
Rather than encoding dynamic visual scenes with a sequence of still images, acquired at a fixed frame rate,
event cameras generate data in the form of a sparse and asynchronous events stream. 
Each event is represented by a tuple $(x,y,p,t)$ corresponding to an illuminance 
change by a fixed relative amount, at pixel location $(x,y)$ and time $t$, with the polarity $p\in\{0,1\}$ indicating whether the illuminance was increasing or decreasing. Fig.~\ref{fig:intro} shows examples of data from an event camera in a driving scenario.

Since the camera does not rely on a global clock, 
but each pixel independently emits 
an event as soon as it detects an illuminance change,
the events stream has a very high temporal resolution, 
typically of the order of microseconds~\cite{Lichtsteiner2008}.
Moreover, due to a logarithmic pixel response characteristic, event cameras have a large dynamic range (often exceeding $120dB$)~\cite{finateu20205}.
Thanks to these properties, event cameras are well suited for applications in which standard frame cameras are affected by
motion blur, pixel saturation, and high latency. 

Despite the remarkable properties of event cameras,
we are still at the dawn of event-based vision and 
their adoption in real systems is currently limited.
This implies scarce availability of algorithms, 
datasets, and tools to manipulate and process events.
Additionally, 
most of the available datasets have 
limited spatial resolution or they are not labeled, 
reducing the range of possible applications~\cite{de2020large,gallego2019event}.

To overcome these limitations, several works have focused on the reconstruction of gray-level information from an event stream~\cite{kim2008simultaneous,bardow2016simultaneous,munda2018real,rebecq2019high}. This approach is appealing since the reconstructed images can be fed to standard computer vision pipelines, leveraging more than 40 years of computer vision research.
In particular, it was shown~\cite{rebecq2019high} that all information required to reconstruct high-quality images is present in the event data.
However, passing through an intermediate intensity image comes at the price of adding considerable computational cost.
In this work, we show how to build an accurate event-based vision pipeline without the need of gray-level supervision.
%

\begin{figure*}[t]
    \centering
    \includegraphics[width=1\textwidth]{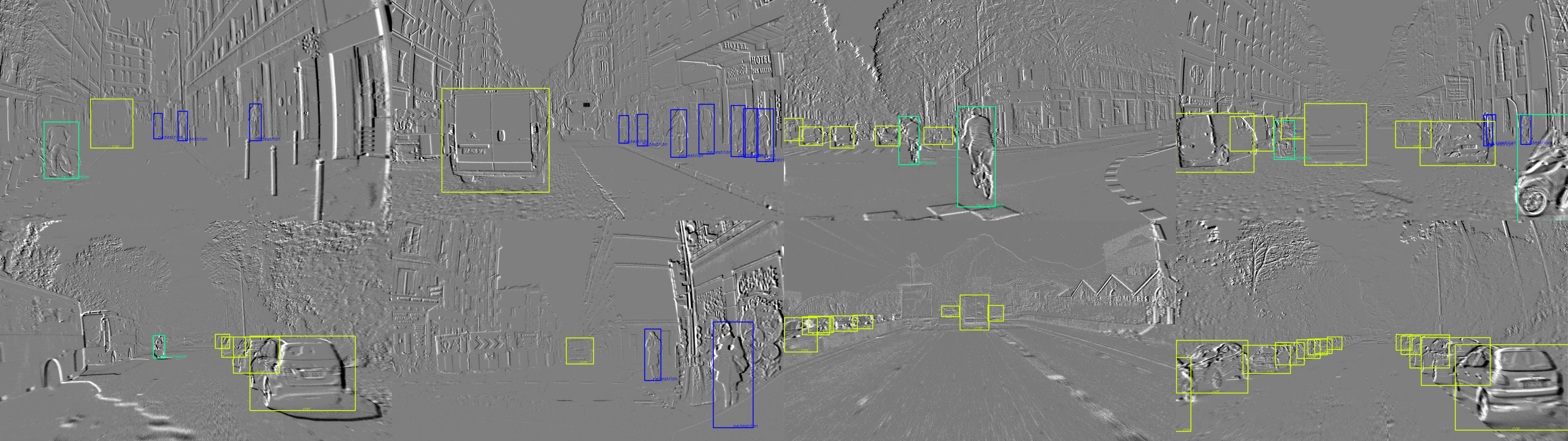}
    \caption{Results of our event-camera detector on examples of the released 1Mpx Automotive Detection Dataset. Our method is able to accurately detect objects for a large variety of appearances, scenarios, and speeds. This makes it the first reliable event-based system on a large-scale vision task. Detected cars, pedestrians and two-wheelers are shown in yellow, blue and cyan boxes respectively.
    All figures in this work are best seen in electronic form.}
     \vspace{-4mm}
\label{fig:intro}
\end{figure*}

\comment{
\begin{figure}[t]
    \centering
    \includegraphics[width=1\textwidth]{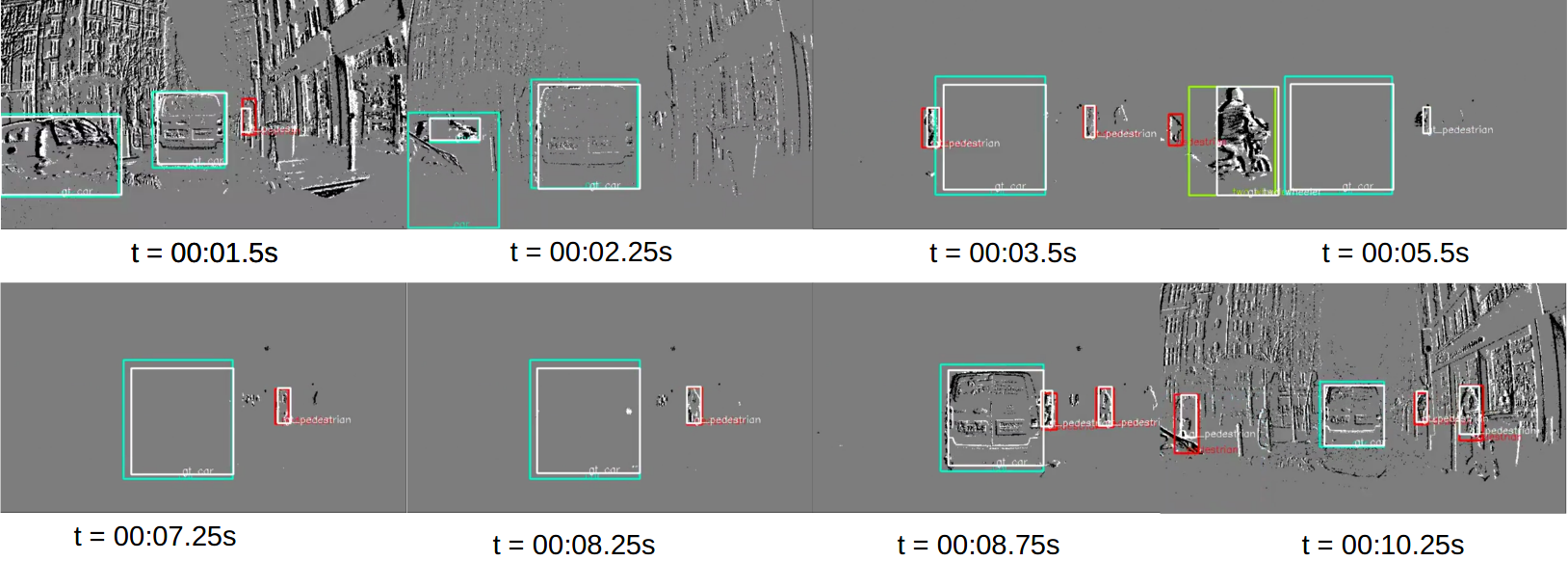}
    \caption{Detecting objects with an event camera is an open computer vision problem. This is in part due to the lack of 
		large labeled datasets and effective machine learning architectures. One of the main challenges for event-based object detection 
		is to keep object persistence over time, even when no events are generated by the objects due to small motion.  
		As shown in the images above, our method is able to reliably detect objects over time also when they stop moving, 
		this thanks to a novel recurrent neural network architecture. We release and evaluate our approach on a
        new high resolution detection dataset, showing state-of-the-art performance and without the need of reconstructing 
        gray-level images.
		In the images above, we show input events accumulated over a fixed time interval, with overlayed object bounding boxes:
		White boxes correspond to ground truth, while cyan, red and green boxes are detections of cars, pedestrains and two-wheelers respectively, 
		returned by our method.
    \amosSrmk{Move this in results/method section and put here teaser of results. Times are made-up for now. Find an example video where another object passes in front of a detected box with no events}}
    \label{fig:intro}
\end{figure}
}
%
We target the problem of object detection in automotive scenarios, which is 
characterized by important objects dynamics and extreme lighting conditions. 
We make the following contributions to the field: 
First, we acquire and release the first large scale dataset for event-based 
object detection, with a high resolution (1280$\times$720) event camera~\cite{finateu20205}. 
We also define a fully automated labeling protocol, enabling fast and cheap dataset 
generation for event cameras. The dataset we release contains more than 
14 hours of driving recording, acquired in a large variety of scenarios. We also provide more 
than 25 million bounding boxes of cars, pedestrians and two-wheelers, labeled at 60Hz. 

Our second contribution is the introduction of a novel architecture for 
event-based object detection together with a new temporal consistency loss.
Recurrent layers are the core building block of our architecture, they introduce a fundamental memory mechanism 
needed to reach high accuracy with event data. At the same time, the temporal consistency loss helps to obtain more precise localization over time.
Fig.~\ref{fig:intro} shows some detections returned by our method on the released dataset.
We show that directly predicting the object locations is more efficient and more accurate than applying a detector on the gray-level images reconstructed with a state-of-the-art method~\cite{rebecq2019high}.
In particular, since we do not impose any bias coming from intensity image supervision,
we let the system learn the relevant features for the given task, which do not necessarily correspond to gray-level values.

Finally, we run extensive experiments on ours and another dataset 
for which gray-level images are also available, showing 
comparable accuracy to standard frame-based detectors and improved state-of-the-art results for event-based detection.
To the best of our knowledge, 
this is the first work showing an event-based system with on par 
performance to a frame-based one on a large vision task.

\section{Related Work}
\label{sec:related_work}
%
%
\vspace{-2mm}
Several machine learning architectures have been proposed for event cameras ~\cite{wang2019space,sekikawa2019eventnet,messikommer2020event}.
Some of these methods, such as Spiking Neural Networks~\cite{kasabov2013dynamic,lee2016training,shrestha2018slayer,tavanaei2019deep}, exploit the sparsity of the data and can be applied event by event, to preserve the temporal resolution of the events stream~\cite{lee2016training,cannici2019asynchronous,sironi2018hats,neil2016phased}. 
However, efficiently applying these methods to inputs with large event rate remains difficult. For these reasons, their efficacy has mainly been demonstrated on low-resolution classification tasks.


Alternative approaches map the events stream 
to a dense representation~\cite{alonso2019ev,maqueda2018event,zhu2019unsupervised,rebecq2019high}.
Once this representation is computed, 
it can be used as input to standard architectures.
Even if these methods lose some of the event's 
temporal resolution, they gain in terms of accuracy and scalability~\cite{zhu2018ev,maqueda2018event}.


Recently, the authors of~\cite{rebecq2019high} showed how to use a recurrent UNet~\cite{ronneberger2015u} to reconstruct high-quality gray-level images from event data. 
The results obtained with this method show the richness of the information contained in the events. 
However, reconstructing a gray-level image before applying a detection algorithm adds a further computational step, 
which is less efficient and less accurate than directly using the events, as we will show in our experiments.

Very few other works have focused directly on the task of event-based object detection. 
In~\cite{cannici2019asynchronous}, the authors propose a sparse convolutional network inspired by the YOLO network~\cite{redmon2016you}.
While in~\cite{li2017adaptive}, 
temporally pooled binary images from the event camera are fed to a faster-RCNN~\cite{ren2015faster}.
However, these methods have only been tested on simple sequences, with a few moving objects on
a static background. As we will see, feed-forward architectures are less accurate 
in more general scenarios. 

The lack of works on event-based object detection is also related to the scarce availability of large benchmarked datasets.
%
Despite the increasing effort of the community~\cite{orchard2015converting,amir2017low,binas2017ddd17,zhu2018multivehicle},
very few datasets provide ground-truth for object detection.
The authors of~\cite{miao2019neuromorphic} provide a pedestrian detection dataset. However, it is composed of only 12 sequences of 30 seconds.
Simulation~\cite{rebecq2018esim,gehrig2019video} is an alternative way to obtain large datasets.
Unfortunately, existing simulators use too simplified hardware models to accurately reproduce all the characteristics of event cameras.
Recently~\cite{de2020large} released an automotive 
dataset for detection. However, it is acquired with a low-resolution QVGA event camera 
and it contains low frequency labels ($\leq4$Hz). We believe instead that high-spatial resolution and
high-labeling frequency are crucial to properly evaluate an automotive detection pipeline. 




\section{Event-based Object Detection}
\label{sec:object_detection}
In this section, we first formalize the problem of object detection with an event camera,
then we introduce our method and the architecture used in our experiments.
\vspace{-2mm}
\subsection{Problem Formulation}
\label{subsec:formulation}
Let $\textbf{E} = \{e_i =(x_i,y_i,p_i,t_i)\}_{i\in\mathbb{N}}$ be an input sequence of events, with 
$x_i \in [0,M]$ and $y_i \in [0,N]$ the spatial coordinates of the event, 
$p_i\in\{0,1\}$ the event's polarity and $t_i\in [0,\infty)$ its timestamp. 
We characterize objects by a set of bounding boxes 
$\textbf{B} = \{b^*_j = (x_j,y_j,w_j,h_j,l_j,t_j)\}_{j\in\mathbb{N}}$,
where, $(x_j,y_j)$ are the coordinates of the top left corner of the bounding box,
$w_j,h_j$ its width and height, $l_j\in\{0,\ldots,\mathrm{L}\}$ the label object class, 
and $t_j$ the time at which the object 
is present in the scene. 
    
A general event-based detector is given by a function $\mathcal{D}$, mapping $\textbf{E}$ to 
$\textbf{B} = \mathcal{D}(\textbf{E})$.
Since we want our system to work in real time, we will assume that the output of a detector 
at time $t$ will only depend on the past, i.e.\ on events generated before $t$: 
$\mathcal{D}(\textbf{E}) = \{D(\{e_i\}_{t_i<t})\}_{t>=0}$,
where $D(\{e_i\}_{t_i<t})$ outputs 
bounding boxes at time $t$. In this work, we want to learn $D$.
    
Applying the detector $D$ at every incoming event is too expensive and often not required by the final 
applications, since the apparent motion of objects in the scene is typically much slower than the pixels response time. 
For this reason, we only apply the detector at fixed time intervals of size $\Delta t$:
\begin{equation}
\mathcal{D}(\textbf{E}) \approx \{D(\{e_i\}_{t_i<t_k})\}_{k\in\mathbb{N}}, 
\label{eq:general_det}
\end{equation}
with $t_k = k\Delta t$\footnote{
Our method can also be applied on a fixed number of events.
For clarity, we only describe the fixed $\Delta t$ case.
}.
However, a function $D$ working an all past events $\{e_i\}_{t_i<t_k}$ for every $k$, 
would be computationally intractable, since the number of input events would indefinitely increase over time. 

A solution would be to consider at each step $k$, only the events in the interval 
$[t_{k-1}, t_k)$, as it is done for example in~\cite{zhu2018ev,maqueda2018event} for other event-based tasks. 
However, as we will see in Sec.~\ref{sec:experiments}, this approach leads to poor results for object detection.
This is mainly due to two reasons: first, it is hard to choose a single $\Delta t$ (or a fixed number of events) working for objects having very different speeds and sizes, such as cars and pedestrians. Secondly, since events contain only relative change information, 
an event-based object detector must keep a memory of the past. 
In fact, when the apparent motion of an object is zero, it does not generate events anymore. 
Tracking objects using hard-coded rules is generally not accurate for edge cases such as reflections, moving shadows or object deformations. 

For these reasons, we decide to learn a memory mechanism  end-to-end, directly from the input events.
%
To use past events information while keeping computational cost tractable, 
we choose $D$ such that 
\begin{equation}
\mathcal{D}(\textbf{E}) \approx \{D(\{e_i\}_{t_i\in[t_{k-1}, t_k)},\textbf{h}_{k-1})\}_{k\in\mathbb{N}},
\end{equation}
where $\textbf{h}_{k-1}$ is an internal state of our model encoding past information at time $t_{k-1}$. For each $k$, we define $\textbf{h}_k$ by a recursive formula $\textbf{h}_k= F(\{e_i\}_{t_i\in[t_{k-1}, t_k)},\textbf{h}_{k-1})$, with $\textbf{h}_0 = \textbf{0}$.
In the next sections, we describe the recurrent neural network architecture we propose to learn $D$ and $F$.
%
\begin{figure}[t]
    \centering
    \includegraphics[width=1\textwidth]{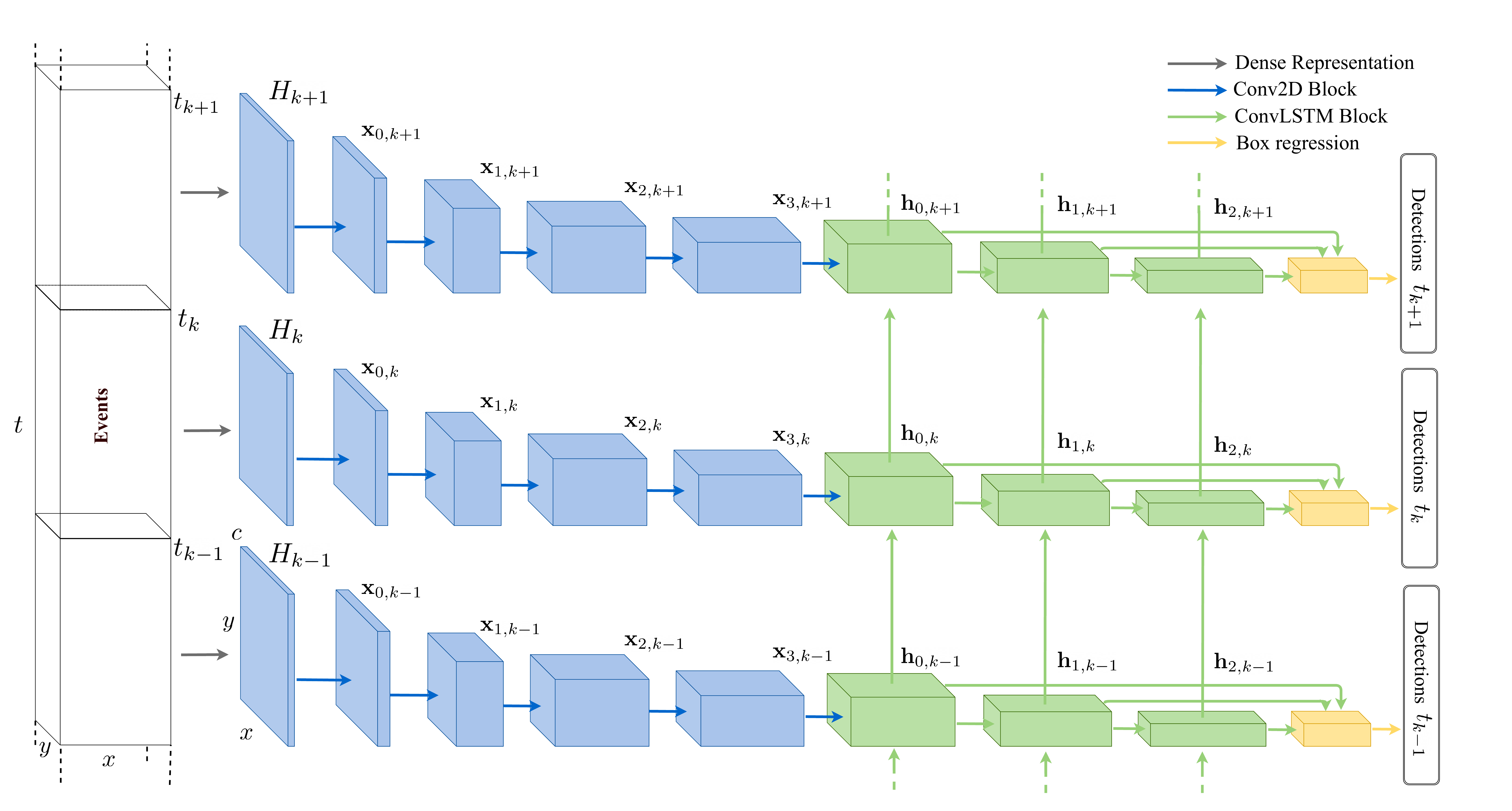}
    \caption{Overview of the proposed architecture. Input events are used to build a tensor map $H_k$ at every time step $t_k$. Feed-forward convolutional layers extract low-level features from $H_k$. Then, ConvLSTM layers extract high-level spatio-temporal patterns. Finally, multiscale features from the recurrent layers are passed to the output layers, to predict bounding box locations and class. 
    Thanks to the memory of the ConvLSTM layers, temporal information is accumulated and preserved over time, allowing robust detections even when objects stop generating events in input.
    }
    \label{fig:method_overview}
    \vspace{-4mm}
\end{figure}

\subsection{Method}
\label{sec:method}
%
%
%
%
%
In this section, we describe the recurrent architecture we use to learn the detector $D$. 
In order to apply our model, we first preprocess the events to build a dense 
representation. More precisely, given input events $\{e_i\}_{t_i\in[t_{k-1}, t_k)}$ 
    we compute a tensor map $H_k\in \mathbb{R}^{\mathrm{C}\times \mathrm{M}\times \mathrm{N}}$, with $\mathrm{C}$ the number of channels. 
We denote $H_k = H$ in the following. Our method is not constrained to a particular $H$ (cfr. Sec.~\ref{subsec:ablation_study}). 
 

To extract relevant features from the spatial component of the events, $H$ is fed as input to a convolutional neural network 
~\cite{krizhevsky2012imagenet,liu2016ssd}. In particular, we use
Squeeze-and-Excitation layers~\cite{hu2017squeezeandexcitation}, 
as they performed better in our experiments.
%
%
%
In addition, we want our architecture to contain a memory state to accumulate 
meaningful features over time and to remember the presence of objects, even when 
they stop generating events. 
For this, we use ConvLSTM layers~\cite{xingjian2015convolutional}, which have been 
successfully used to extract spatio-temporal information from 
data~\cite{finn2016unsupervised,liu2018mobile}.
Our model first uses $K_{f}$ feed-forward convolutional layers to extract high-level semantic features that are then fed to the remaining $K_{r}$ ConvLSTM layers
(cfr. Fig.~\ref{fig:method_overview}). 
This is to reduce the computational complexity and memory footprint of the method due to 
recurrent layers operating on large feature maps, and more importantly to avoid the recurrent 
layers to model the dynamics of low-level features that is not necessary for the given task.
We denote this first part of the network \textit{feature extractor}.

The output of the feature extractor is fed to a bounding box  \textit{regression head}. In this work, we use Single Shot Detectors (SSD)~\cite{liu2016ssd}, since they are a good compromise between accuracy and computational time. However, our feature extractor could be used in combination with other detector families, such as two-stage detectors. 
%
%
Since we want to extract objects for a large range of scales, we feed features at 
different resolutions to the regression head. In practice, we use the feature map from
each of the recurrent layers.
%
A schematic representation of our architecture is provided in
Fig.~\ref{fig:method_overview}.
As typically done for object detection, to train the parameters of our network, we optimize a loss function composed of a regression term $\mathcal{L}_r$ for the box coordinates and 
a classification term $\mathcal{L}_c$ for the class.
We use smooth $l1$ loss~\cite{liu2016ssd} $\mathcal{L}_{s}$ for regression and the softmax focal loss~\cite{lin2017focal} for classification.
More precisely, for a set of $\mathrm{J}$ ground-truth bounding boxes at time $t_k$, we encode their coordinates in a tensor $B^*$ of size $(\mathrm{J}\cdot\mathrm{R},4)$, as done in~\cite{liu2016ssd}, where $\mathrm{R}$ is the number of default boxes of the regression head matching a ground-truth box.
Let $(B,p)$ be the output of the regression head, with $B$ the tensor encoding the prediction for the above $\mathrm{R}$ default boxes and $p$ the class probability distribution for all default boxes.
Then, the regression and classification terms of the loss are:
\begin{equation}
\mathcal{L}_r=
\mathcal{L}_{s}(B,B^*), \quad  \mathcal{L}_c = -(1-p_l)^\gamma \log p_l,
\label{eq:l1loss}
\end{equation}
where $p_l$ is the probability of the correct class $l$.
We set the constant $\gamma$ to 2 and also adapt the unbalanced biases for softmax logits in the spirit of~\cite{lin2017focal}.
\begin{figure*}[t]
    \centering
      \begin{tabular}{cc}
        \includegraphics[width=0.45\textwidth]{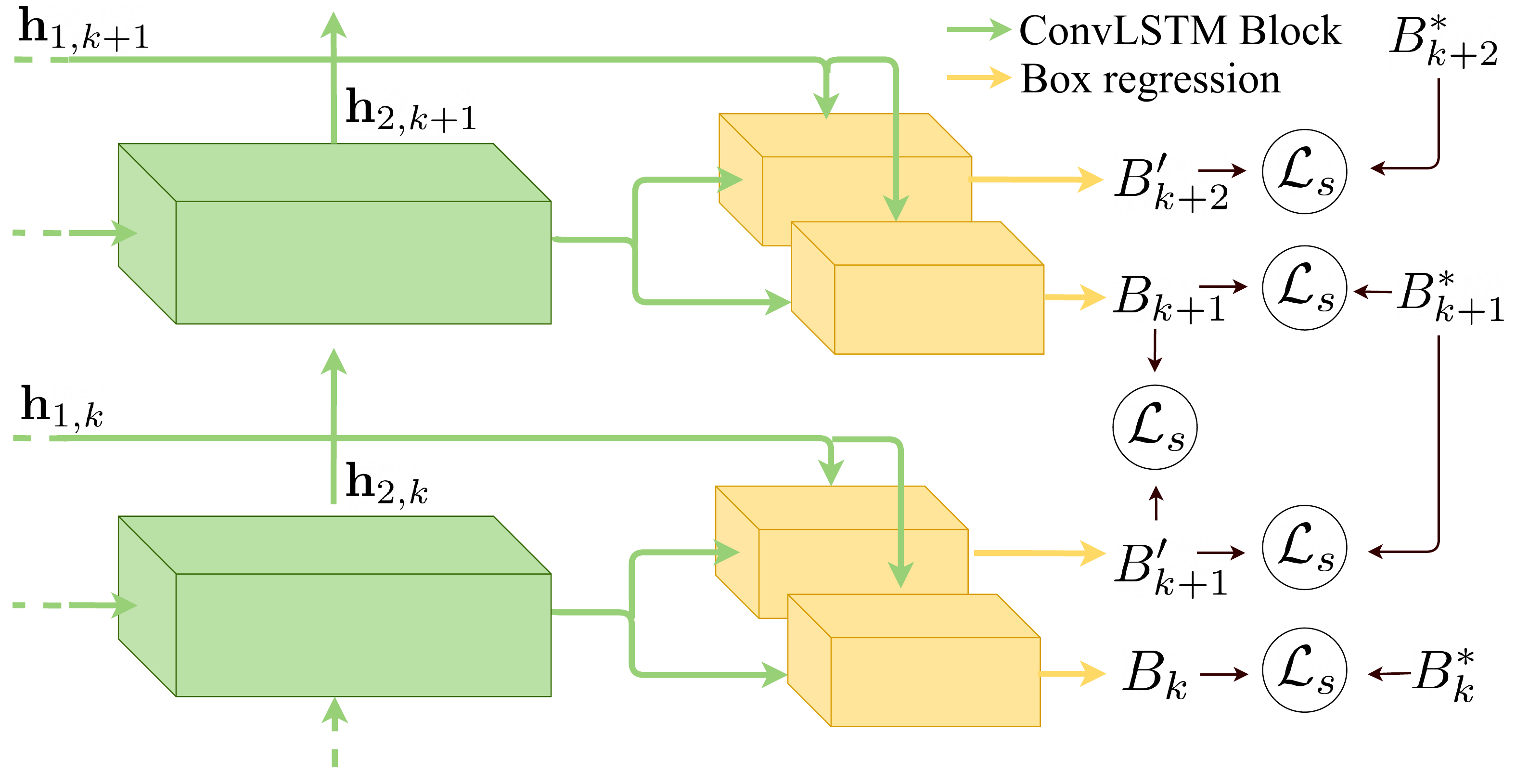}&
        \includegraphics[width=0.45\textwidth]{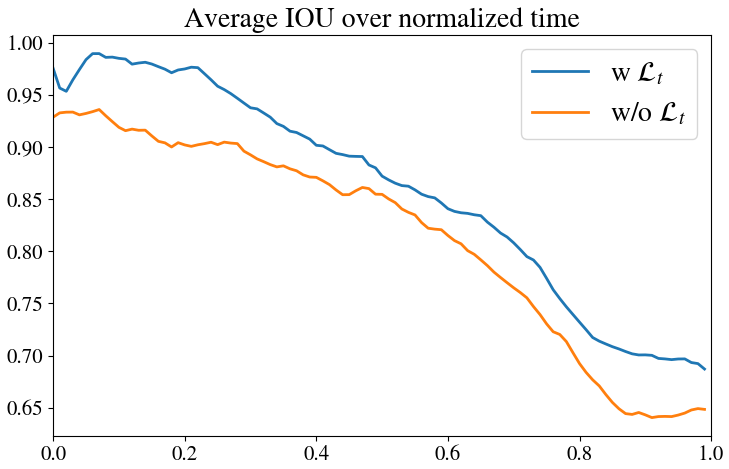}\\
        \vspace{-2mm}
        (a)&
        (b)
	\end{tabular}
    \caption{(a) Detail of the box regression heads. In order to regularize temporally our network, we introduce a secondary regression head, predicting, at time $t_k$, the boxes $B'_{k+1}$ for time $t_{k+1}$. We impose predictions corresponding to the same time step to be consistent. (b) IoU between ground truth tracks and predicted boxes over time. The consistency loss helps obtaining more precise boxes.}
    \vspace{-5mm}
\label{fig:consistency}
\end{figure*}
\subsubsection{Dual Regression Head and Temporal Consistency Loss}
\label{subsec:consistency}
To have temporally consistent detections, we would like the 
internal states of the recurrent layers to learn high-level features that are stable over long periods of time.
Even if ConvLSTM can, to some extent, learn slow-changing representations, we further improve detections consistency by introducing an auxiliary loss and an additional regression head trained to predict bounding boxes one time-step into the future. 
The idea is inspired by unsupervised learning methods such as word2vec~\cite{word2vec} and CPC~\cite{cpc}, 
which constraint the latent representation to preserve some domain structure. 
In our case, given that features are shared for both heads, we argue that this has the 
additional effect of inducing representations that account for object motion, something
which is currently used as regularization, but would require further analysis that goes
beyond the scope of the current work.


%
%
Given the input tensor $H_k$ computed in the time interval $[t_{k-1},t_k)$, the two regression heads will output bounding boxes
$B_k$ and $B'_{k+1}$, trying to match ground truth $B^{*}_k$ and $B^{*}_{k+1}$ respectively.
This dual regression mechanism is shown in Fig.~\ref{fig:consistency}(a). 

To train the two regression heads, we add to our loss an auxiliary regression term between $B'_{k+1}$ and $B^{*}_{k+1}$. 
This term, when applied at every time step $k$, indirectly constraints the output $B'_{k}$'s of the second head to be close to the predictions $B_{k}$'s of the first 
head at the next time step, cfr. Fig.~\ref{fig:consistency}(a). 
However, since the two heads are independent, they could converge to different
solutions. Therefore, we further regularize training by adding another loss term
explicitly imposing $B'_{k}$ to be close to $B_{k}$. %
In summary, the auxiliary loss is:
\begin{equation}
     \mathcal{L}_{t} = \mathcal{L}_{s}(B'_{k+1},B^{*}_{k+1}) + \mathcal{L}_{s}(B'_{k},B_{k}).
     \label{eq:consistency_total}
\end{equation}
Then, the final loss we use during training is given by $\mathcal{L} = \mathcal{L}_c + \mathcal{L}_r + \mathcal{L}_t$.
We minimize it during training using truncated backpropagation through time~\cite{werbos1990backpropagation}.

\section{The 1 Megapixel Automotive Detection Dataset}
\label{sec:datasets}
\vspace{-2mm}
In this section, we describe an automated protocol to generate datasets for event cameras.
We apply this protocol to generate the detection dataset used in our experiments. 
However, our approach can be easily adapted 
to other computer vision tasks, such as face detection and 3D pose estimation. 
\vspace{-1mm}
\paragraph{Setup and Fully Automated Labeling Protocol}
%
The key component to obtaining automated labels is to do recordings with an event camera and a standard RGB camera
side by side. Labels are first extracted from the RGB camera and then transferred to the event camera pixel coordinates by using a geometric transformation.
%
In our work, we used the 1 megapixel event camera of~\cite{finateu20205} and a GoPro Hero6.
The two cameras were fixed on a rigid mount side by side, 
as close as possible to minimize parallax errors.
For both cameras, we used a large field of view:
~110 degrees for the event camera and ~120 degrees for the RGB camera. The video stream of the RGB camera is recorded at 4 megapixels and 60fps.
%
%
Once data are acquired from the setup, we perform the following label transfer:
%
1. Synchronize the time of the event and frame cameras; 2. Extract bounding boxes from the frame camera images; 3. Map the bounding box coordinates from the frame camera to the event camera.
The bounding boxes from the RGB video stream are obtained using a commercial automotive detector, outperforming freely available ones.
The software returns labels corresponding to pedestrians, two-wheelers, and cars. 
%
%
The time synchronization can be done using a physical connection between the cameras.
However, since this is not always possible, we also propose in the supplementary material an algorithmic way of synchronizing them.
%
%
%
Once the 2 signals are synchronized temporally, we need to find a geometric transformation mapping pixels from the RGB camera to the event camera.
Since the distance between the two cameras is small,
the spatial registration can be approximated by a homography. 
Both time synchronization and homography estimation can introduce some noise in the labels. 
Nonetheless, we observed time synchronization errors smaller than the discretization step $\Delta t$ we use, and that
the homography assumption is good enough for our case, since objects encountered in automotive scenarios are relatively far compared to the cameras baseline.
We discuss more in depth failure cases of the labeling protocol in Sec.~\ref{subsec:failure_cases}.
More details can be also be found in the supplementary material.

\vspace{-3mm}
\paragraph{Recordings and Dataset Statistics}
Once the labeling protocol is defined, we can easily collect and label a large amount of data.
To this end, we mounted the event and frame cameras behind 
the windshield of a car. 
%
We asked a driver to drive in a variety of scenarios, including city,
highway, countryside, small villages, and suburbs. 
The data collection was conducted over several months, 
with a large variety of lighting and weather conditions during daytime.
%
At the end of the recording campaign, a total of 14.65 hours was obtained. 
We split them in 11.19 hours for training, 2.21 hours for validation, and 2.25 hours for testing.
%
The total number of bounding boxes is 25M.
More statistics can be found in the supplementary material, together with examples from the dataset.
%
To the best of our knowledge, the presented event-based dataset is the largest in terms of labels and classes.
Moreover, it is the only available high-resolution detection dataset for event cameras\footnote{Dataset available at: \href{https://www.prophesee.ai/category/dataset/}{prophesee.ai/category/dataset/}}.



%




\vspace{-3mm}
\section{Experiments}
\label{sec:experiments}
\vspace{-2mm}
In this section, we first evaluate the importance of the main components of our method in an ablation study. Then, we compare it against state-of-the-art detectors.
We consider the COCO metrics~\cite{lin2014microsoft}
and we report COCO mAP, as is it widely used for evaluating detection algorithms. Even if this metric is designed for frame-based data, we explain in the supplementary material how we extend it to event data.
Since labeling was done with a 4 Mpx camera, but the input events have lower resolution, in all our experiments, 
we filter boxes with diagonal smaller than 60 pixels. 
All networks are trained for 20 epochs using ADAM~\cite{kingma2014adam} and learning rate $0.0002$ with exponential decay of $0.98$ every epoch. 
We then select the best model on the validation set and apply it to the test set to report the final mAP.
\footnote{Evaluation code at \href{https://github.com/prophesee-ai/prophesee-automotive-dataset-toolbox}{github.com/prophesee-ai/prophesee-automotive-dataset-toolbox}
}
\vspace{-2mm}
\subsection{Ablation Study}
\label{subsec:ablation_study}
As explained in Sec.~\ref{sec:method}, our network can take different representations as input. Here we compare the commonly used Histograms of events~\cite{moeys2016steering,maqueda2018event}, Time Surfaces~\cite{lagorce2016hots}, and Event Volumes~\cite{zhu2019unsupervised}. The results are given in Tab.~\ref{tab:ablation_study}. We see that Event Volume performs the best. Time Surface is $2\%$ points less accurate than Event Volume, but more accurate than simple Histograms. 
We notice that we could also learn the input representation $H$ together with the network. For example by combining it with~\cite{cannici2020matrix,tulyakov2019learning,gehrig2019end}. However, for efficiency reasons, we decided to use a predefined representation and introduce the memory mechanism in the deeper layers of the network, rather than at the pixel level.

In a second set of experiments, we show the importance of the internal memory of our network.
To do so, we train while constraining the internal state of the recurrent layers to zero. 
As we can see in Tab.~\ref{tab:ablation_study}, the performance drops by 12\%, showing that the memory of the network is fundamental to reach good accuracy.
%
Finally, we show the advantage of using the loss $\mathcal{L}_{t}$ of Sec.~\ref{subsec:consistency}.
When training with this term, mAP increases by  $2\%$, and COCO  $\text{mAP}_{75}$ increases by  $4\%$, Tab.~\ref{tab:ablation_study}. This shows the advantage of using $\mathcal{L}_{t}$, especially with regard to box precision. 
In order to better understand the impact of the loss $\mathcal{L}_{t}$ on the results, we compute the Intersection over Union (IoU) between 1000 ground truth tracks from the validation set and the predicted boxes. We normalize the duration of the tracks to obtain the average IoU. As shown in Fig.~\ref{fig:consistency}(b), with $\mathcal{L}_t$, IoU is higher for all tracks duration.   

\begin{figure*}[t]
    \centering
     \includegraphics[width=1\textwidth]{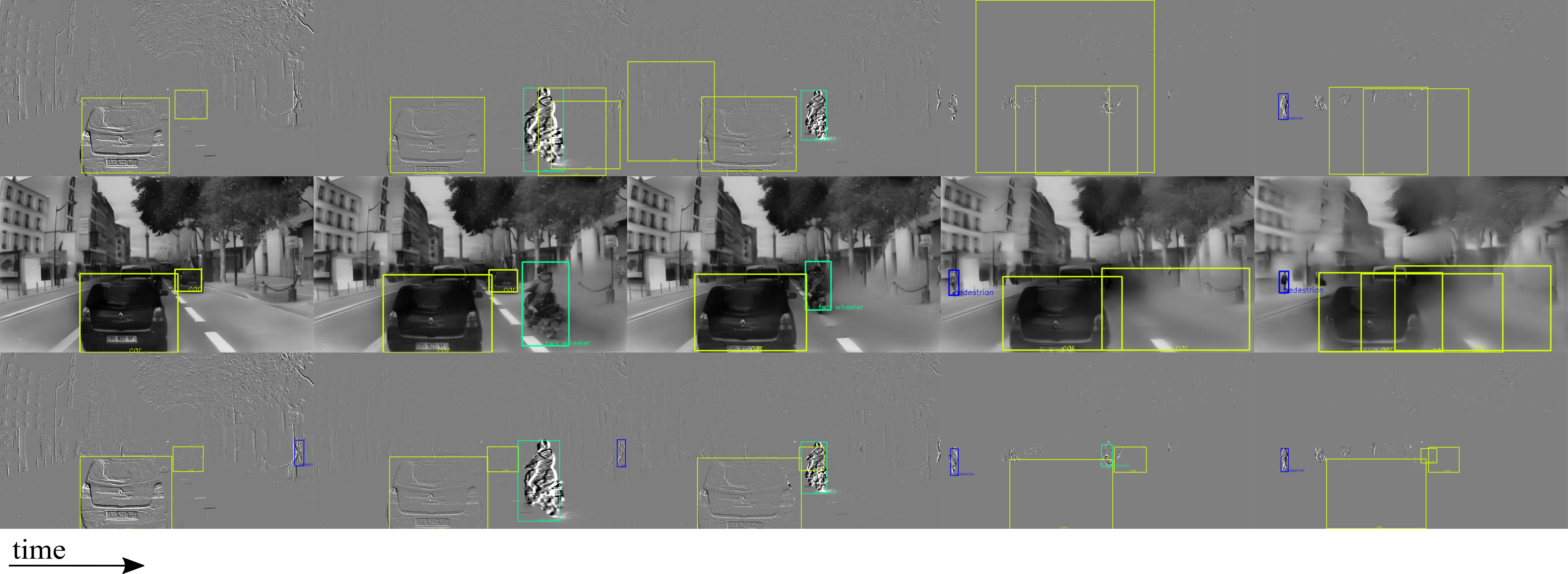}
    \caption{Detections on a 1 Mpx Dataset sequence. 
    From top to bottom: \textbf{Events-RetinaNet}, \textbf{E2Vid-RetinaNet} (with the input reconstructed images), and our method \textbf{RED}. 
    Thanks to the memory representation learned by our network, 
    \textbf{RED} can detect objects even when they stop generating events, as for example the stopped car on the right, even when occluded by the motorbike.}
\label{fig:gen4_results}
\end{figure*}
\begin{table}[htbp]
    \caption{Ablation study on the 1Mpx  Dataset. \textbf{Left}: mAP for different input representations (without consistency loss). \textbf{Right}: mAP and $\text{mAP}_{75}$ without some components of our method, with Event Volume as input. "w/o memory" means forcing the internal state $\textbf{h}_k$ to be zero, for all recurrent layers.}
    \vspace{1mm}
    \begin{subtable}{.5\linewidth}
      \centering
		\tabcolsep=0.11cm
		\begin{tabular}{@{}ccc@{}}
			\toprule
			 Histogram & Time Surface & Event Volume\\             
			\hline
		     0.37  & 0.39   &   0.41\\
			\bottomrule
		\end{tabular}
    \end{subtable}%
    \begin{subtable}{.5\linewidth}
      \centering
		\tabcolsep=0.11cm
		\begin{tabular}{@{}ccc@{}}
			\toprule
			w/o memory & w/o $\mathcal{L}_t$ loss & memory + $\mathcal{L}_t$ loss\\             
			\hline
			0.29 | 0.26 &  0.41 | 0.40   & 0.43 | 0.44     \\    
			\bottomrule
		\end{tabular}
    \end{subtable} 
    \label{tab:ablation_study}
\end{table}
\subsection{Comparison with the State-of-the-art}
\label{subsec:results}
We now compare our method with the state-of-the-art on the 1 Mpx Detection Dataset and the Gen1 Detection Dataset~\cite{de2020large}, which is another automotive dataset acquired with a QVGA event camera~\cite{posch2010qvga}.

We denote our approach \textbf{RED} for Recurrent Event-camera Detector.
For these experiments, we consider Event Volumes of 50ms. 
%
Since there are not many available algorithms for event-based detection,
we use as a baseline a feed-forward architecture applied to the same input representation as ours,
thus emulating the approach of~\cite{miao2019neuromorphic} and~\cite{li2017adaptive}. 
We considered several architectures, leading to similar results. We report here those of RetinaNet~\cite{lin2017focal} with ResNet50~\cite{he2016deep} backbone and a feature pyramid scheme, since it gave the best results. We refer to this approach as \textbf{Events-RetinaNet}.
Then, we consider the method of~\cite{rebecq2019high}, 
which is currently the best method to reconstruct graylevel images from events and uses a recurrent Unet. For this, we use code and network publicly released by the authors. Then, we train the RetinaNet detector on these images.
We refer to this approach as \textbf{E2Vid-RetinaNet}. 
For all methods, before passing the input to the first convolutional layer of the detector, 
input height and width are downsampled by a factor 2.
For the Gen1 Detection Dataset, we report also results available from the literature~\cite{messikommer2020event}.

Finally, since the 1 Mpx Dataset was recorded together with a RGB camera, we can train a frame-based detector on these images. Since events do not contain color information, we first convert the RGB images to grayscale. Moreover, to have the same level of noise in the labels due to the automated labeling, we map frame camera pixels to the same resolution and FOV as the event camera.
In this way, we can have an estimation 
of how a grayscale detector would perform on our dataset.
Similarly, since the Gen1 Dataset was acquired using an event camera providing gray levels, we could run the RetinaNet detector on them.
%
%
We refer to this approach as \textbf{Gray-RetinaNet}.
%

The results we obtain are given in Tab.~\ref{tab:results}. We also report the number of parameters of the networks and the methods runtime, including both events preprocessing and detector inference, on a i7 CPU at 2.70GHz and a GTX980 GPU.  
Qualitative results are provided in Fig.~\ref{fig:gen4_results} and in the supplementary material.
From Fig.~\ref{fig:gen4_results}, we see in particular that our model continues detecting the car even when it does not generate events.
While \textbf{Events-RetinaNet} becomes unstable and \textbf{E2Vid-RetinaNet} begins oversmoothing the image, and thus loses the detection. 
As we can see from Tab.~\ref{tab:results}, our method outperforms all the other event-based ones by a large margin. On the 1Mpx dataset the images reconstructed by~\cite{rebecq2019high}  are of good quality and therefore \textbf{E2Vid-RetinaNet} is the second best method, even if 18\% points behind ours. Instead, on the Gen1 Dataset, the model of~\cite{rebecq2019high} does not generalize well and images are of lower quality. Therefore, on this dataset, \textbf{Events-RetinaNet} scores better. 
Our method reaches the same mAP as \textbf{Gray-RetinaNet} on the 1Mpx Dataset, making it the first event-camera detector with comparable precision to that of commonly used frame-camera detectors. 
If we also consider color, the RetinaNet mAP increases to 0.56, confirming that color information is usefull to increase the accuracy. Our method could benefit from color if acquired by the sensor, such as in~\cite{li2015design}.
On the Gen1 Dataset our method performs slightly worse, this is due to the higher level of noise of the QVGA sensor and also because the labels lower frequency makes training a recurrent model more difficult.
Finally, we observe that our method has less parameters than the others, it runs realtime and, on the 1Mpx Dataset, it is 21x faster than \textbf{E2Vid-RetinaNet}, which reconstructs intensity images.
%
\begin{table}[htpb]
\vspace{-3mm}
	\caption{Evaluation on the two automotive detection datasets.}
	\begin{center}
		\tabcolsep=0.11cm
		\begin{tabular}{l ccc |cc}
			\toprule
			& \multicolumn{3}{c}{1Mpx Detection Dataset} & \multicolumn{2}{c}{Gen1 Detection Dataset}\\
			\hline
			                                                                      & {mAP}  & runtime (ms) & params (M)  & {mAP}  & runtime (ms) \\
			\hline
			\textbf{MatrixLSTM}~\cite{cannici2020matrix}                          & -      & -       & -           &   0.31$^*$     & - \\
			\textbf{SparseConv}~\cite{messikommer2020event}                       & -      & -       & -           &   0.15     & - \\
			\textbf{Events-RetinaNet}                                             & 0.18   & 44.05    & 32.8       &   0.34     & 18.29 \\
			\textbf{E2Vid-RetinaNet}                                              & 0.25   & 840.66   & 43.5       &   0.27     & 263.28 \\
			\textbf{RED} (ours)                                                   & \textbf{0.43}   & \textbf{39.33}    & \textbf{24.1}       &   0.40     &\textbf{16.70} \\
			\hline
			\textbf{Gray-RetinaNet}                                               & \textbf{0.43}   & 41.43    & 32.8        &  \textbf{0.44}    & 17.35 \\
			\bottomrule
			\multicolumn{3}{l}{\small{$^*$ Provided by the authors, using a pretrained YOLOv3.}}
		\end{tabular}
	\end{center}
 	\vspace{-4mm}
	\label{tab:results}
\end{table}
%
\begin{figure*}[t]
    \centering
      \begin{tabular}{@{}c@{}c@{}c@{}c}
        \includegraphics[width=0.25\textwidth]{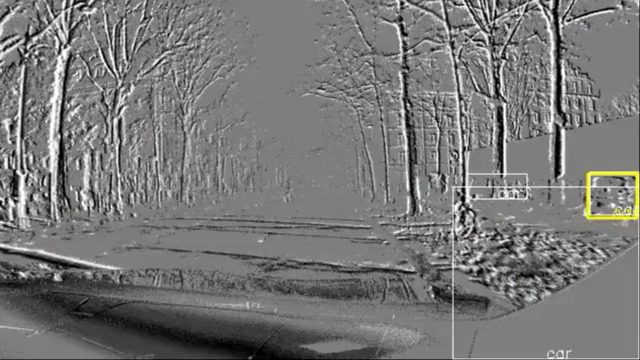}&
        \includegraphics[width=0.25\textwidth]{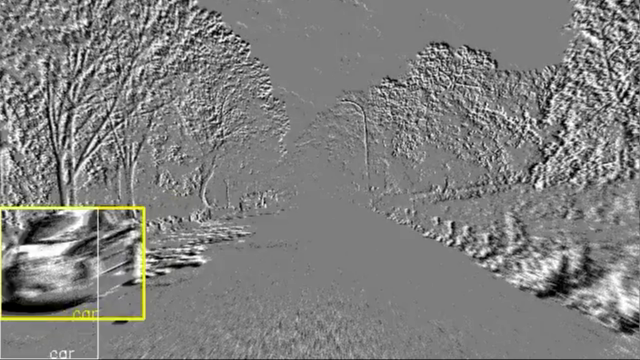}&
        \includegraphics[width=0.25\textwidth]{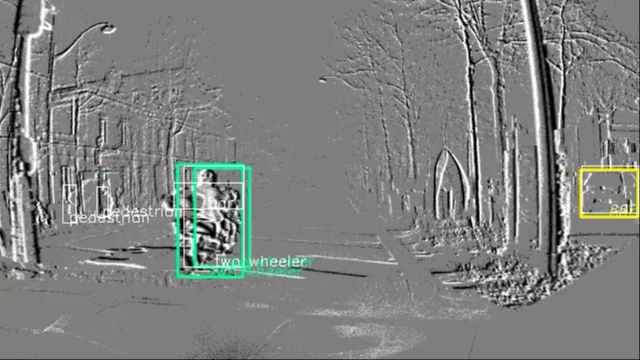}&
        \includegraphics[width=0.25\textwidth]{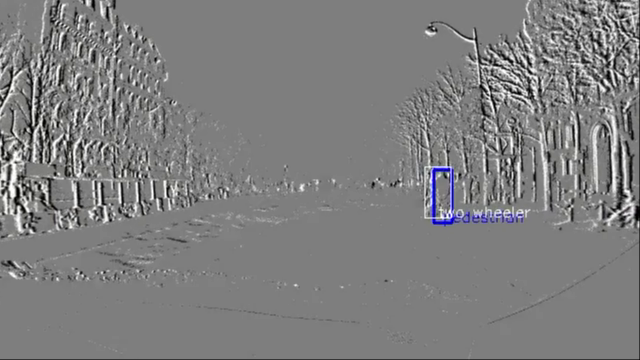}
	\end{tabular}
    \caption{Labeling and detector failure cases. White boxes correspond to labels, colored boxes to detections. From left to right: Labels outlier, labels misalignment, double detection, label swap.}
    \vspace{-2mm}
\label{fig:failure_cases}
\end{figure*}
\vspace{-1mm}
\subsection{Failure Cases}
\label{subsec:failure_cases}
In this section, we discuss some of the failure cases of our network and of the automated labeling protocol of Sec.~\ref{sec:datasets}. In Fig.~\ref{fig:failure_cases}, we show the results of our detector together with the ground truth on some example sequences. In the ground truth we observe two types of errors: geometric errors and semantic errors. Geometric errors, such as misalignment between objects and bounding boxes are due to an imprecise temporal and spatial registration between the event and the frame cameras. Semantic errors, such as label swaps or erroneous boxes, are due to wrong detections of the frame based software used for labeling. Our detector can correct some of the geometric errors, if the errors in the box position are uniformly distribuited around the objects.
Semantic labels are harder to correct, and an outlier robust loss might be beneficial during training. In addition, we observe that our detector can produce double detections and is less accurate on small objects. 

\subsection{Generalization to Night Recordings and Other Event Cameras}
\label{subsec:night_results} 
We now study the generalization capabilities of our detector.
First, we focus on applying our detector, trained with daylight data only, on night recordings.
Since event cameras are invariant to absolute illuminance levels, an event-based detector should 
generalize better than a frame-based one.
%
To test this, we apply \textbf{RED} and \textbf{Gray-RetinaNet} detectors on
new recorded night sequences, captured using the event camera of 
Sec.~\ref{sec:datasets} and a HDR automotive camera.
We stress the fact that these networks have been trained exclusively on daylight data. Since the frame-based labeling software of Sec.~\ref{sec:datasets} is not accurate enough for night data, we report qualitative results in Fig.~\ref{fig:night} and in the appendix.
It can be observed that the accuracy of \textbf{Gray-RetinaNet} drops considerably.
This is due to the very different lighting and the higher level of motion blur
inherently present in night sequences. On the contrary, our method performs
well also in these conditions. 

\begin{figure*}[t]
    \centering
      \begin{tabular}{@{}c@{}c@{}c@{}c}
        \includegraphics[width=0.25\textwidth]{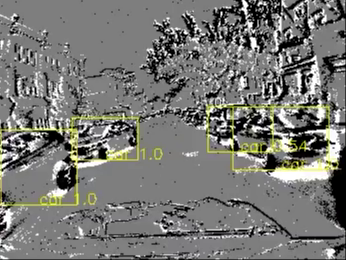}&
        \includegraphics[width=0.25\textwidth]{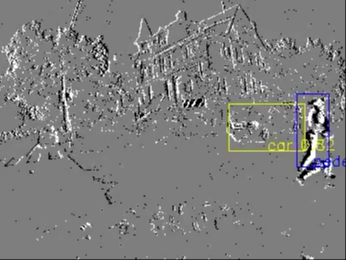}&
        \includegraphics[width=0.25\textwidth]{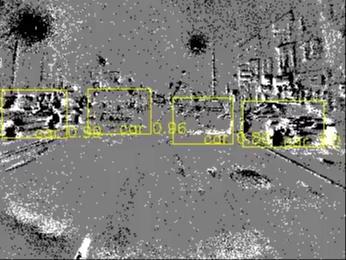}&
        \includegraphics[width=0.25\textwidth]{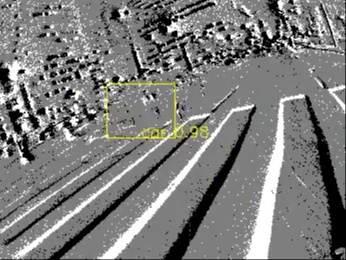}
	\end{tabular}
    \caption{\textbf{RED} detector trained on ATIS data and applied to DAVIS sequences. Even if our model was trained on a different camera, it generalizes to other sensors, points of view and light conditions.}
    \vspace{-4mm}
\label{fig:davis_det}
\end{figure*}

In a second experiment, we test the generalization capability of our network when using different camera types as input.
Since there is no available 
dataset with object detection labels, 
we report qualitative results on the MVSEC dataset~\cite{zhu2018multivehicle}, which is an automotive dataset acquired with a DAVIS-346 camera.
For this purpose we use the model trained on the Gen1 Dataset, since it was acquired with an ATIS camera with similar resolution as the DAVIS. From Fig.~\ref{fig:davis_det}, we see that even if the model was trained on a different camera, it generalizes well also on the DAVIS sequences.
\begin{figure*}[t]
    \centering
      \begin{tabular}{@{}c@{}c@{}c}
        \includegraphics[width=0.33\textwidth]{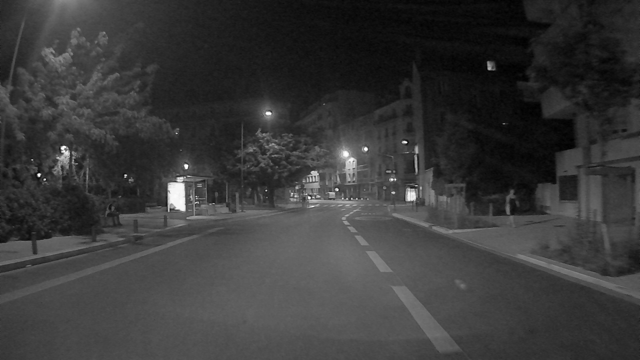}&
        \includegraphics[width=0.33\textwidth]{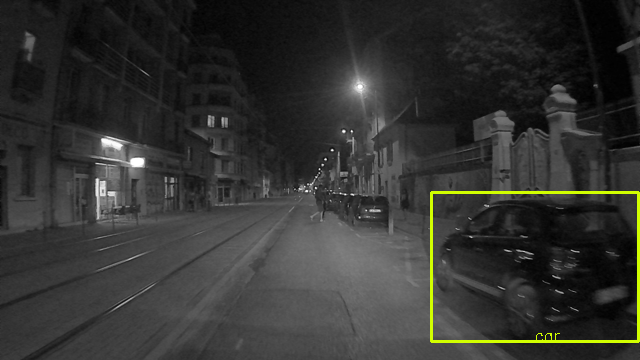}&
        \includegraphics[width=0.33\textwidth]{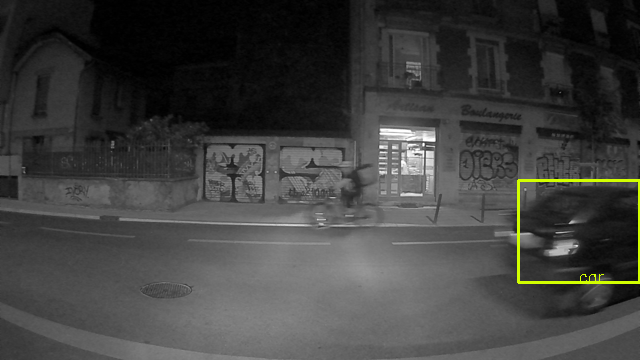}\\[-3.3pt]
         \includegraphics[width=0.33\textwidth]{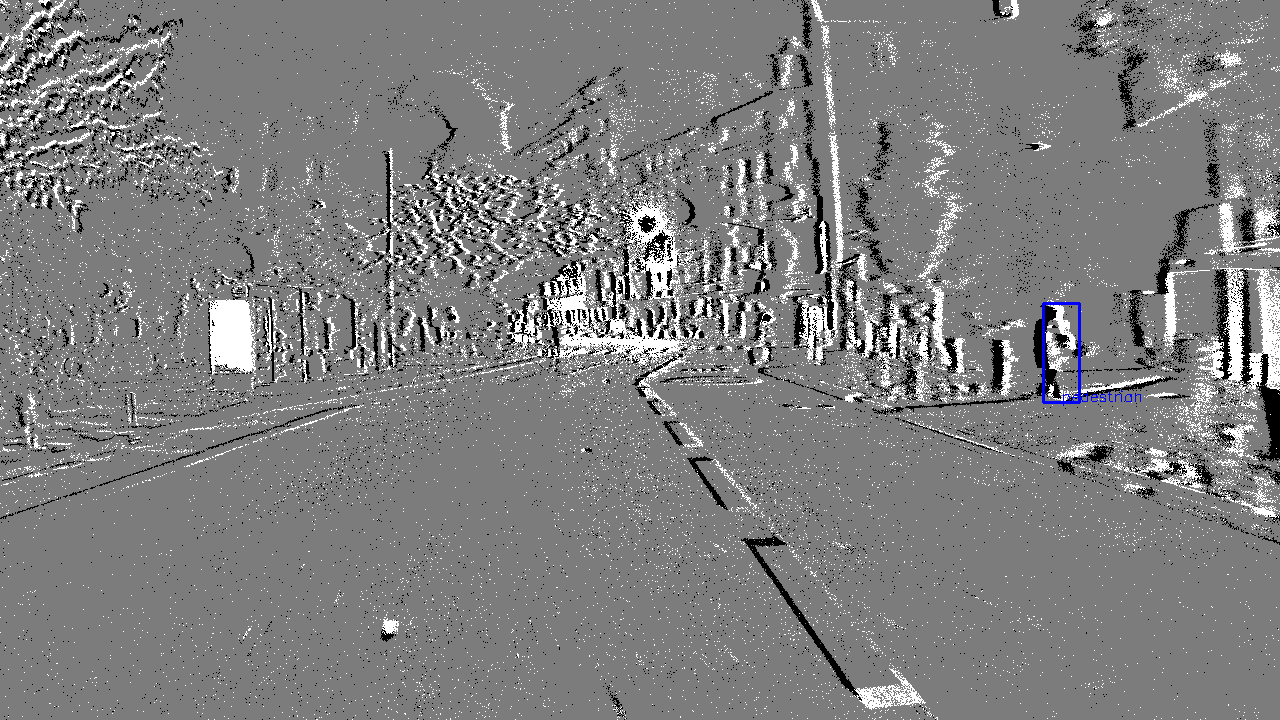}&
        \includegraphics[width=0.33\textwidth]{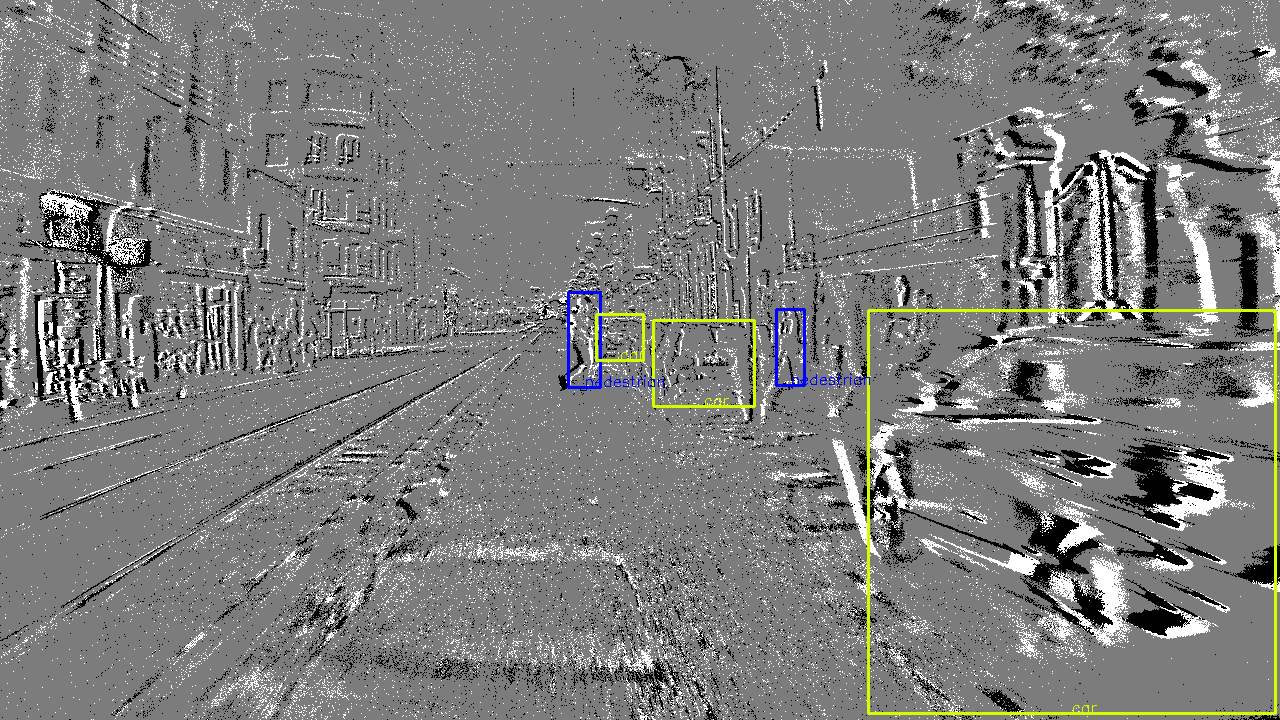}&
        \includegraphics[width=0.33\textwidth]{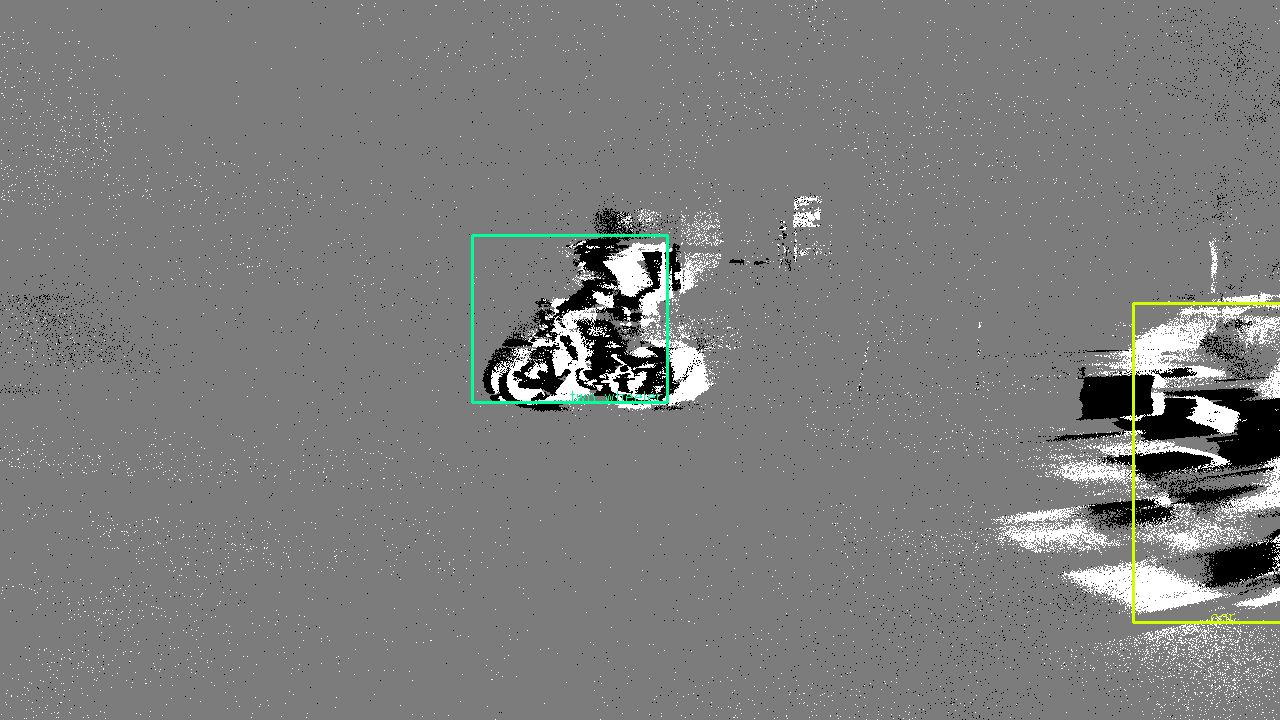}
	\end{tabular}
    \caption{\textbf{Top}: \textbf{Gray-Retinanet} applied to night recordings of a HDR automotive camera. \textbf{Bottom}: Our detector \textbf{RED} 
    applied to recordings of a 1 Mpx event camera of the same scene.
    The detectors were trained  on day light data.
    \textbf{Gray-Retinanet} does not generalize well on night images.
    In contrast, \textbf{RED} generalizes on night sequences because event data is invariant to absolute illuminance levels.}
    \vspace{-2mm}
\label{fig:night}
\end{figure*}

\vspace{-2mm}

\section{Conclusion}
\label{sec:conclusion}
We presented a high-resolution event-based detection dataset and a real-time recurrent neural network architecture which can detect objects from event cameras with the same accuracy as mainstream gray-level detectors.
We showed it is possible to consistently detect objects over time without the need for an intermediate gray-level image reconstruction.
However, our method still needs to pass through a dense event representation. This means that our method does not take advantage of the input data sparsity. In the future, we plan to exploit the sparsity of
the events to further reduce computational cost and latency. This could be done for example by adapting our method to run on neuromorphic hardware~\cite{akopyan2015truenorth,davies2018loihi}.

\section*{Broader Impact}
The integration of an event-based object detection pipeline in real-world applications could positively impact several aspects of existing systems. First, the camera's high temporal resolution would allow faster reaction time and be more robust in situations where standard cameras suffer from motion blur or high latency. Secondly, they could also improve performance in HDR or low light scenes. Both these aspects are essential to increase the safety of driving assistance solutions or autonomous vehicles~\cite{combs2019automated}.  
Similarly, these characteristics could be useful in applications where there is an interaction between humans and robots (e.g., in a production line or in a warehouse).
Finally, the adoption of similar pipelines in other contexts, like the Internet of Things, could reduce the power consumption and the data storage of existing systems~\cite{martin2020literature}.

Although, as demonstrated in~\cite{rebecq2019high}, the possibility to reconstruct intensity images from events stream could create privacy issues, the proposed method allows better privacy management. Encoding events in not human-readable structures and not requiring to have an image-like representation prevents the 
easy use of the recorded data for purposes different than those defined by the original algorithm, 
and it limits the possibility to identify people, vehicles or places.

Further advances in event-based processing and neuromorphic architectures might also open the future to a new class of extremely low-power and low-latency artificial intelligence systems~\cite{rajendran2019low}. In a world where power-hungry deep learning techniques are becoming a commodity, and at the same time, environmental concerns are increasingly pressuring our way of life, neuromorphic systems could be an essential component of a sustainable society~\cite{Eliasmith2020scaling}. 

Concerning possible negative outcomes, since our method relies on training data, it will leverage the bias and the limitations contained in it. Similarly, since it relies on deep learning architectures, it might be deceived by adversarial attacks. 
To mitigate these consequences, several methods have been recently proposed to de-bias deep learning models and make them more robust  to adversarial examples
~\cite{zhang2018mitigating,lecuyer2019certified,jha2019attribution,svoboda2018peernets}. 
A failure of the system 
might cause dangerous incidents and have severe consequences on people and facilities  
~\cite{tsugawa2006trends}.
Similarly, its integration in a fully autonomous
vehicle, poses the ethical question of replacing the human morale in the so called Trolley Problem~\cite{maxmen2018self}. Moreover, autonomous vehicles may impact the careers of millions of people~\cite{balakrishnan2017goldman}.

Finally, we think it is essential to be aware that the event-based perception and similar detection systems could be exploited to harm people and threaten human rights. For example, developing modified versions of this algorithm for mass surveillance~\cite{feldstein2019global} or military applications~\cite{cummings2017artificial,Knight2019military}.



\begin{ack}

    We would like to thank Davide Migliore for the organization and the support inside the company during the whole duration of paper preparation. We would also like to thank Cécile Pignon for acquiring and reviewing most of the sequences of 1 Megapixel dataset. Finally, thank to Matteo Matteucci and Marco Cannici from Politecnico di Milano for fruitful discussions and for providing the MatrixLSTM results of Table 2.
    
    This publication and the related work was performed in the scope of the ES3CAP research project, under the Bpifrance Invest for the Future Program (\textit{Programme d’Investissements d’Avenir} — PIA).
    This work was also funded in part by the EU NEOTERIC H2020-ICT-2019-2 project 871330.

\end{ack}

\newpage
\appendix
\section*{Supplementary Material}

In this supplementary material, we report additional details which did not fit in our submission due to space limitations.
In Sec.~\ref{sup:datasets}, we provide more statistics about the released 1 Mpx Automotive Detection Dataset and details about the automated labeling protocol, introduced in Sec. 4 of the main submission. 
Then, in Sec.~\ref{subsec:evaluation_methodology}, we explain how to adapt the COCO metric for an event-based object detection task. In Sec.~\ref{sec:losses} and Sec.~\ref{sec:representation}, we formally define the losses and the input representations used in our experiments. Finally, in Sec.~\ref{sec:architecture} we provide the architecture used in our experiments,

Attached to this appendix, we also provide a video with the results of our method on sample sequences from the two detection datasets used in our experiments and a comparison with the frame-based detector on night recordings.

\section{The 1 Megapixel Automotive Detection Dataset}
\label{sup:datasets}

As described in our submission, we build our dataset thanks to a fully automated labeling protocol for event cameras. In this section, we provide more statistics about our dataset and more details about the automated labeling protocol.

\begin{figure}[tphb]
    \centering
    \includegraphics[width=1\textwidth]{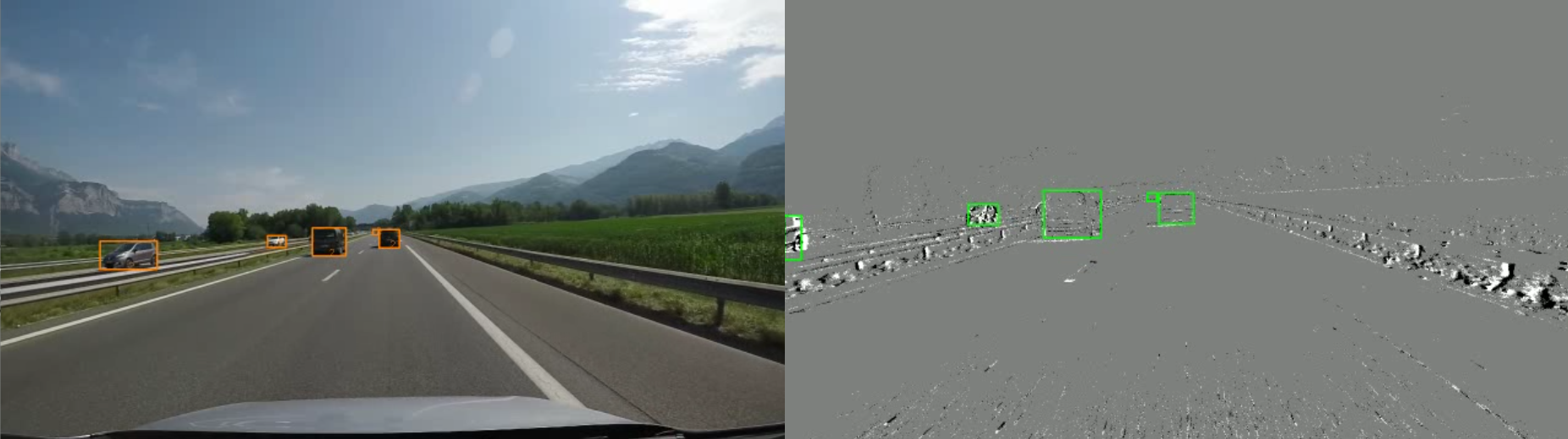}
  
    \includegraphics[width=1\textwidth]{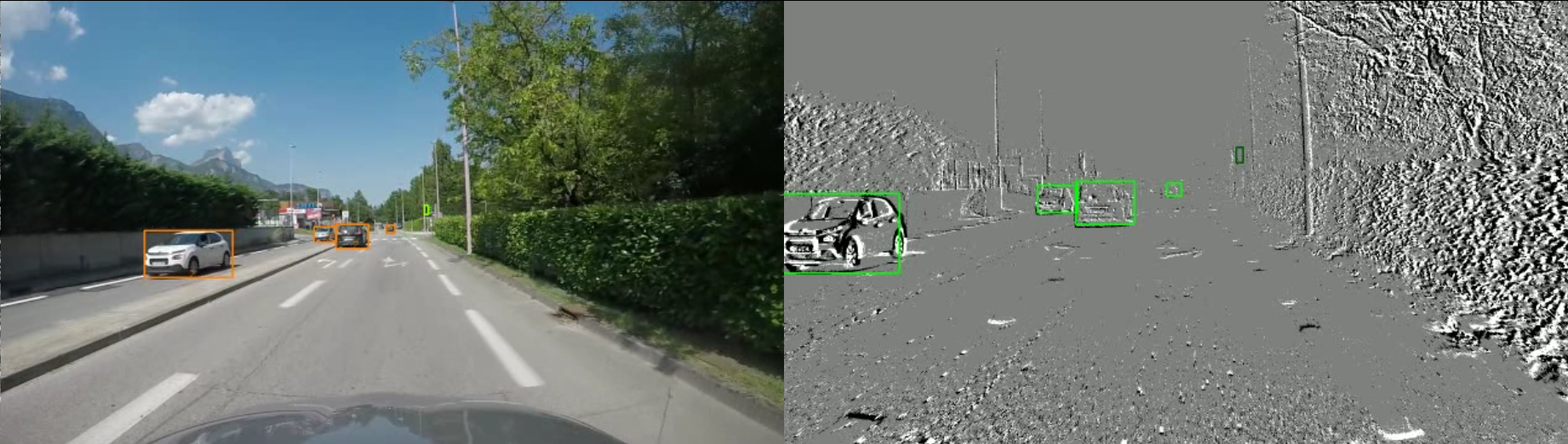}
  
    \includegraphics[width=1\textwidth]{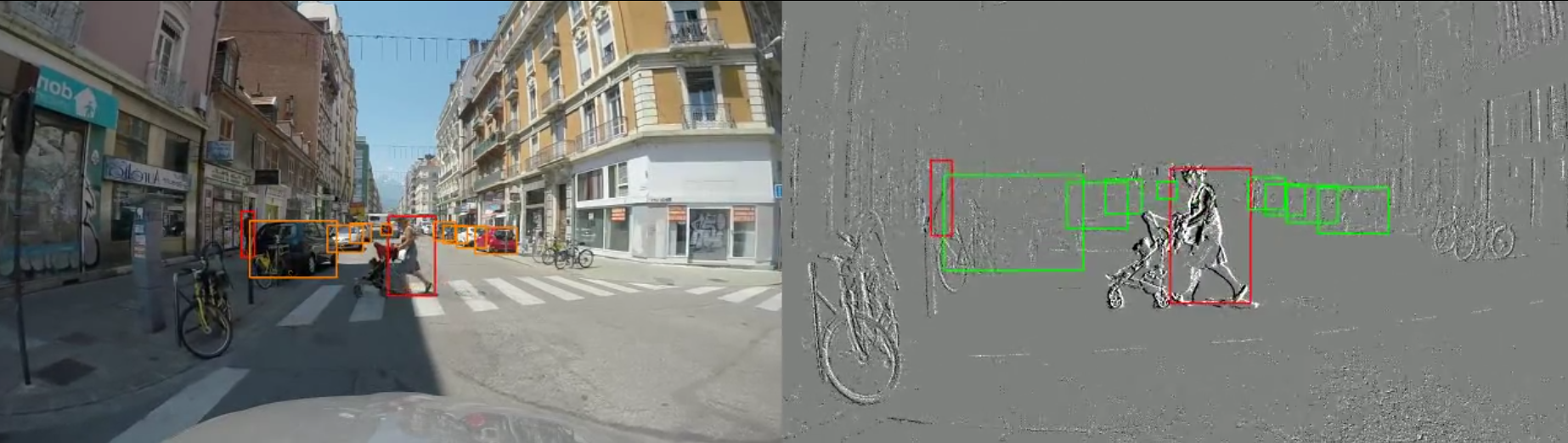}
    \caption{\textbf{Left}: RGB frames used for labelling, with overlaid bounding boxes returned by the automotive labeling software. \textbf{Right}: sample snapshots from the 1Mpx Detection Dataset, with transferred bounding boxes. The bounding boxes are transferred from the RGB camera using the automoated labeling protocol described in Sec.~\ref{sup:datasets}. }
    \label{fig:dataset_samples}
\end{figure}

\paragraph{Dataset Statistics}
The 1Mpx Automotive Detection Dataset is composed of 14.65 hours or recordings of a 1280x720 event camera~\cite{finateu20205}. Data are recorded in different scenarios, including city, with different level of traffic, highway, countryside, small villages and suburbs. 
The data collection campaign was conducted over several months, 
with a large variety of lighting and weather conditions during daytime.
Recordings are labeled with the automated protocol described in the main submission, yielding 25 million bounding boxes of cars, pedestrians and two-wheelers (i.e.\ bikes and motorbikes).
We split the recordings in 11.19 hours for train, 2.21 hours for validation 
and 2.25 hours for test. 
Moreover, to ease training and evaluation each recording is split in 60 seconds chunks, leaving at least one minute between chunks belonging to the different splits.  
Precise statistics are given in Tab.~\ref{tab:statistics}. Sample images form the dataset are given in Fig \ref{fig:dataset_samples}. 
\begin{table}[htpb]
	\caption{Number of labels per class on the 1 Mpx Automotive Detection Dataset.}
	\begin{center}
		\tabcolsep=0.11cm
		\begin{tabular}{lccc}
			\toprule
			 & \textbf{Car} & \textbf{Pedestrian} & \textbf{Two-wheeler}\\
			\hline
            Train & 11,567,763 & 6,022,634 & 802,707\\
            Validation & 2,339,095 & 919,596 & 138,934\\
            Test & 2,396,966 & 1,487,702  & 171,428\\
            \hline
            Total & 16,303,824 & 8,429,932 & 1,113,069\\
			\bottomrule
		\end{tabular}
	\end{center}
	\label{tab:statistics}
\end{table}

\paragraph{Time Synchronization and Spatial Registration}

The labeling protocol consists in transferring labels from RGB camera data to event camera data, recorded side by side.
In the following, we describe an algorithmic way to synchronize the two cameras and to obtain the homography mapping RGB pixels coordinates to event camera pixel coordinates.

Time synchronization is obtained by using zero-normalized cross-correlation (ZNCC) applied 
on one-dimensional statistics extracted from the two signals. 
The 1D signals extracted from the event-based camera are the sum and the standard 
deviation of the number of events per pixel in a time slice of $1/60s$. 
The 1D signals extracted from the RGB camera are the sum and standard deviation 
of the intensity value of the absolute difference of consecutive gray-level frames. 
For each pair of signals (e.g., sum of events and sum of intensity of frame difference) 
the ZNCC is computed and the highest value is the estimate of temporal difference between 
the event-based camera and the RGB camera. All estimates are finally averaged using the median.


%


To estimate the homography, we extract feature points from the two data streams.
For the RGB stream, we simply extract FAST~\cite{rosten2006machine} points.
For the event camera, feature points are obtained by extracting Harris points from
histogram of events in a time slice of length $1/60s$.
Once feature points are obtained, we can fit an homography using standard methods, 
such as RANSAC~\cite{fischler1981random}.

An alternative solution to estimate the homography consists in minimizing the squared difference between two images (pixel by pixel) where one image is the histogram of events and the other image is the difference of gray level frames. Such loss is then minimized using gradient descent.

\section{Evaluation Methodology}
\label{subsec:evaluation_methodology}
In this section, we describe how we adapt the COCO metric protocol to work with event data.

We first notice that for the datasets we consider, ground-truth labels are given by
a set of bounding boxes at fixed frequency. 
As a consequence, if the detections returned by an algorithm have the same frequency as the ground truth, accuracy computation is equivalent to evaluating a frame-based detection algorithm.

For the more general case in which detections have a different rate than the ground truth,
or when detections are returned in an asynchronous fashion, evaluation can be reduced to the previous case.
This is achieved by restricting evaluation to only timestamps for which both detections and ground-truth information are available. If necessary one can add a small time tolerance on the detections timestamps.

For both datasets we consider, the ground-truth is annotated starting from gray-level images.
For this reason, there might be some ground-truth bounding boxes at the beginning of a sequence
for which no events have been generated yet (such as a static object in front of the sensor when the recording car is stopped). Since it would be impossible for
an event-based algorithm to predict the presence of an object in this particular condition, 
we decide to ignore such boxes during training and evaluation in the first 0.5 seconds of each sequence. 
More precisely, no loss is computed during training for $0.5s$ and the validation is skipped for the same amount of time.

Similarly, since the frame-based camera of Sec.~\ref{sup:datasets}, used for annotation has larger resolution,
very far objects are not clearly distinguishable in the event camera.
For this reason, we ignore bounding boxes with diagonal size smaller then 60 pixels or smaller then 20 pixel in width or height, 
during training and evaluation.

\subsection{Losses}
\label{sec:losses}
In this section, we formally define the losses used in the paper.
\subsection{Softmax Focal Loss}
For the classification term $\mathcal{L}_c$ of our loss, we consider the Focal Loss~\cite{lin2017focal}. In the original paper, the authors of ~\cite{lin2017focal} used a one-versus-all setting with sigmoid. Instead, in our experiments, we obtain better results using a softmax setting.

The focal loss allows to concentrate on hard examples more efficiently in highly unbalanced cases such as semantic segmentation or object detection. To do so, the individual terms of the cross-entropy loss are weighted with $(1-p_l)^\gamma$ where $p_l$ is the probability of the correct class. It has the effect that a big error term will count exponentially more, compensating for the rare classes. One problem arises at the beginning of training, where most probabilities are random. Therefore, most of negative examples will still overwhelm the loss for a lot of iterations. As a solution, the authors of \cite{lin2017focal} provide an efficient biased initialization to increase probability of classifying everything as negative. In their one-versus-all setting, each logit predicts a class versus background (the probability is computed with sigmoid), meaning there is no competition among logits for a single object. 
They initialize biases of all logits $s_l$ 
with $-\log{\frac{p_x}{1 - p_x}} $, where $p_x$ is set to $0.01$. In this way, after applying the sigmoid, the initial probability to predict the background class is around $1-p_x=0.99$ (it would be exactly $0.99$ if the weights were zero). 

We adapt the same principle to initialize the probability in the case of a softmax classification.
We set the logit biases of non-background classes $s_x=0$ and the logit bias for background $s_{\emptyset}=\log ({C \frac{p_{\emptyset}}{1 - p_{\emptyset}}}) $ with $p_{\emptyset}=0.99$ and $C$ the number of object classes (excluding background). Applying the softmax leads to a probability for background class close to $p_{\emptyset}=0.99$. 
To understand this let us assume that the probability of a (non-background) class $p_x$ is equal for each class $p_x=\frac{1-p_{\emptyset}}{C}$ and from the softmax formula we know that $p_x=\frac{e^{s_x}}{e^{s_{\emptyset}} + C e^{s_x} }$. Developing the following gives rise to difference of logits: $s_{\emptyset} - s_x$ = $\log ({C \frac{p_{\emptyset}}{1 - p_{\emptyset}}}) $. Therefore we use this formula for bias of background and set the biases of other classes $s_x$ to zero. 

\subsection{Smooth l1 loss}
For the regression loss $\mathcal{L}_r$ and the auxiliary loss $\mathcal{L}_t$, we consider the smooth $l1$ loss~\cite{liu2016ssd}, denoted as $\mathcal{L}_{s}$.
The loss $\mathcal{L}_{s}(B,B^*)$ applied to a tensor $B$ and ground truth $B^*$  is given by the following:
\begin{align}
\mathcal{L}_s(B,B^*)&=\frac{1}{N}\sum_j
\mathcal{L}_{s}(B_j,B^*_j) \\
    \mathcal{L}_{s}(B_j,B^*_j) &=  
    \begin{cases}
    |B_j-B^*_j| - \frac{\beta}{2}  & \text{if } |B_j-B^*_j|\geq \beta \\
    \frac{1}{2 \beta} (B_j-B^*_j)^2 & \text{otherwise}
    \end{cases}
\end{align}
where $B_j$ and $B^*_j$ are the elements of $B$ and $B^*$ respectively.
In the experiments, we set $\beta=0.11$.

\section{Input Representation}
\label{sec:representation}

 In this section, we formally define the input tensor representations we consider in the Ablation Study of the main submission, namely Histograms of events~\cite{moeys2016steering,maqueda2018event}, Time Surfaces~\cite{lagorce2016hots}, and Event Volumes~\cite{zhu2019unsupervised}.
In the following we assume we are given an input sequence of events $\{ e_i = (x_i,y_i,p_i,t_i)\}_{i=1}^{I}$ in a time interval of size $\Delta t$.

\subsection{Histogram}
A Histogram of events is simply given by the sum of events per pixels. We also sum the events independently per polarity, by using one channel per polarity. We clamp at maximum value $m$ of events per pixels and then we divide by this maximum value. This leads to a tensor input of size $(2,M,N)$, where a generic element is the following: 
\begin{align}
H_{p,x,y} = \min \left(1,\sum_{i, x_i=x, y_i=y, p_i=p} 1/m\right).
\end{align}
In our experiments, we use $m=20$.

\subsection{Timesurface}
A Timesurface is a 2D snapshot of the latest timestamps of events for a given receptive field. It can be seen as a proxy to the normal optical flow~\cite{benosman2012asynchronous}. If we accumulate event's absolute timestamps on this array, we obtain a buffer containing all latest timestamps per pixel since the beginning of the record. To give less weight to old events, we apply an exponential decay with parameter $\tau_j$ to the timestamps. Thus, assuming for simplicity $t_0=0$, the timesurface $T$ is given by: 

\begin{align}
T_{p,j,x,y} &= \exp\left( \frac{ts_{p,x,y}-\max_{x,y}(ts_{p,x,y})}{\tau_j}\right),\\
ts_{p,x,y} &= \max_{i,x_i=x,y_i=y,p_i=p} t_i .
\end{align}

In order to account for slow and fast motion, we consider two decays $\tau_j$ of 10 and 100 ms. This way we can see both motion gradients in very recent and also moderately far in the past without need of any rolling buffer. The formulation being differentiable,
 we could also learn the set of decay constants $\tau_j$, but we let this for future work.
The timesurface input shape used in the experiments is $(4,M,N)$ where the first dimension refers to the 2 decays for each polarity.

\begin{figure}[t]
    \centering
    \includegraphics[width=1\textwidth]{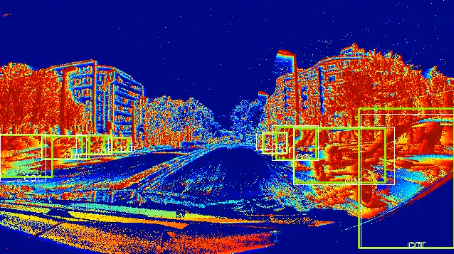}
    \caption{Visualization of a timesurface with a constant decay of 100 ms. Large values are represented with warm colors, small values with cool colors.
    }
    \label{fig:capture_timesurface}
\end{figure}

\subsection{Event Volume}

Event Volumes have been introduced in \cite{zhu2019unsupervised}. In the original work, the contribution from the two polarities is subtracted, here instead we consider each polarity independently. Given the input events $\{ e_i = (x_i,y_i,p_i,t_i)\}_{i=1}^{I}$, the corresponding Event Volume $V$ is given by:
\begin{align}
V_{t,p,x,y} &= \sum_{i, x_i=x, y_i=y, p_i=p} \max(0,1-|t-t_i^*|) \\
t_i^* &= (B-1) \frac{t_i-t_0}{t_I-t_1}
\end{align}
We omit the bilinear kernel for x, y as we restrict to build the volume at maximum resolution. We simply downsample after with a dense operator.  In the experiments, we generate Event Volumes of $B=5$ bins and 2 polarities (for ON and OFF events). The input tensor shape used in the experiment is $(10,M,N)$, where bins and polarity are combined in the first dimension. 


\section{Neural Network Architecture}
\label{sec:architecture}
The proposed neural network \textbf{RED} consists of a feature extractor that is fed to  bounding box regression heads.
The feedforward part of the proposed feature extractor uses Squeeze-Excite Layers, denoted SE and described in Tab.~\ref{tab:architecture2}. The ConvLSTM uses a BatchNorm with Conv Layer for input-to-hidden connection, and a plain Conv Layer for hidden-to-hidden connection. We run input-to-hidden connections (including the BatchNorm layer) in parallel for all timesteps of a batch. Detailed number of layers and parameters for each layer are given in Tab.~\ref{tab:architecture}.

\begin{table}[p]
	\caption{Architecture of the Feature Extractor.}
	\begin{center}
		\tabcolsep=0.11cm
		\begin{tabular}{lcccc}
			\toprule
			 & \textbf{Layer Type} & \textbf{Kernel size} & \textbf{Channels Out} & \textbf{Stride}\\
			\hline
            Layer1 & BNConvReLU & 7 & 32 & 2\\
            Layer2 & Squeeze-Excite & 3 & 64 & 2\\
            Layer3 & Squeeze-Excite & 3 & 64 & 2 \\ 
            Layer4 & ConvLSTM & 3 & 256 & 2\\
            Layer5 & ConvLSTM & 3 & 256 & 2\\
            Layer6 & ConvLSTM & 3 & 256 & 2\\
            Layer7 & ConvLSTM & 3 & 256 & 2\\
            Layer8 & ConvLSTM & 3 & 256 & 2\\
			\bottomrule
		\end{tabular}
	\end{center}
	\label{tab:architecture}
\end{table}

\begin{table}[htpb]
	\caption{Architecture of the Squeeze-Excitation (SE) Layers.}
	\begin{center}
		\tabcolsep=0.11cm
		\begin{tabular}{lcccc}
			\toprule
			 & \textbf{Layer Type} & \textbf{Kernel size} & \textbf{Channels Out} & \textbf{Stride}\\
			\hline
            Layer1 & BNConvReLU & 3 & 256 & 2\\
            Layer2 & BNConvReLU & 3 & 256 & 1\\
            Layer3 & BNConv & 3 & 256 & 2 \\ 
            Layer4 & GlobalAvgeragePooling & MxN & 256 & MxN\\
            Layer5 & DenseReLU & 1 & 64 & 1\\
            Layer6 & DenseSigmoid & 1 & 256 & 1\\
            Layer7 & Elementwise-Multiplication & 1 & 256 & 1\\
            Layer8 & Skip-Sum & 1 & 256 & 1\\
			\bottomrule
		\end{tabular}
	\end{center}
	\label{tab:architecture2}
\end{table}
\clearpage

\small

\bibliographystyle{unsrtnat}
\bibliography{bibliography}
\end{document}